\documentclass{bmvc2k}

\usepackage{cleveref}
\crefname{section}{Sec.}{Secs.}
\Crefname{section}{Section}{Sections}
\Crefname{table}{Table}{Tables}
\crefname{table}{Tab.}{Tabs.}
\Crefname{figure}{Fig.}{Figs.}

\usepackage{xparse}
\usepackage{mathtools}
\DeclarePairedDelimiter{\ceil}{\lceil}{\rceil}
\usepackage{tabularx}
\usepackage{booktabs}
\usepackage{soul}
\usepackage{wrapfig} 
\usepackage{xr}
\usepackage{cuted}%

\usepackage{amsmath}
\usepackage{amssymb}
\usepackage{xr}
\usepackage{pifont}
\usepackage{amsfonts}
\usepackage{graphicx}

\usepackage{subfiles}

\newcommand{\xmark}{\ding{55}}

\title{MILA: Memory-Based Instance-Level Adaptation for Cross-Domain Object Detection}

\addauthor{Onkar Krishna}{onkar.krishna.vb@hitachi.com}{1}
\addauthor{Hiroki Ohashi}{hiroki.ohashi.uo@hitachi.com}{1}
\addauthor{Saptarshi Sinha}{saptarshi.sinha@bristol.ac.uk}{2}

\addinstitution{
 Hitachi Ltd.\\
 Tokyo, Japan
}
\addinstitution{
 University of Bristol\\
 Bristol, UK
}

\runninghead{Onkar, Hiroki, Saptarshi}{MILA}

\def\eg{\emph{e.g}\bmvaOneDot}

\begin{document}
\maketitle
\begin{abstract}
Cross-domain object detection is challenging, and it involves aligning labeled source and unlabeled target domains. 
Previous approaches have used adversarial training to align features at both image-level and instance-level.
At the instance level, finding a suitable source sample that aligns with a target sample is crucial. 
A source sample is considered suitable if it differs from the target sample only in domain, without differences in 
unimportant characteristics such as orientation and color, which can hinder the model's focus on aligning the domain difference. 
However, existing instance-level feature alignment methods struggle to find suitable source instances because their search scope is limited to mini-batches.
Mini-batches are often so small in size that they do not always contain suitable source instances.
The insufficient diversity of mini-batches becomes problematic particularly when the target instances have high intra-class variance.
To address this issue, we propose a memory-based instance-level domain adaptation framework. 
Our method aligns a target instance with the most similar source instance of the same category retrieved from a memory storage. 
Specifically, we introduce a memory module that dynamically stores the pooled features of all labeled source instances, categorized by their labels. Additionally, we introduce a simple yet effective memory retrieval module that retrieves a set of matching memory slots for target instances. 
Our experiments on various domain shift scenarios demonstrate that our approach outperforms existing non-memory-based methods significantly.

%
%

\end{abstract}

\section{Introduction}
\label{sec:intro}
%
Although recent object detection models have achieved success on public datasets~\cite{deng2009imagenet, lin2014microsoft, everingham2010Pascal}, they often suffer from a drop in performance when deployed in real-world use cases due to their inability to automatically generalize to unseen target environments. Retraining models on the target is a possible solution, but it is not viable due to the high cost of annotations.

To address this issue, Unsupervised Domain Adaptation (UDA) leverages knowledge transfer from a labeled source domain to an unlabeled target domain. 
The previous UDA approaches for object detection have focused on aligning the instance-level features extracted from source and target object proposals, which are intermediately generated by the detector model. However, most previous works~\cite{he2019multi, rezaeianaran2021seeking} ignore the category of the instances during the alignment, leading to negative knowledge transfer. 
To solve this, some recent works~\cite{tian2021knowledge, xu2020cross, zhang2021rpn, zheng2020cross} proposed category-to-category (C2C) alignment methods. They employ various techniques to identify instances with matching categories in a sampled mini-batch of source and target images, and then align these instances using either adversarial~\cite{rezaeianaran2021seeking, vs2021mega} or contrastive learning~\cite{zhang2021rpn, xu2020cross, zheng2020cross}. 

\begin{figure*}[t!]
  \includegraphics[width=\textwidth]{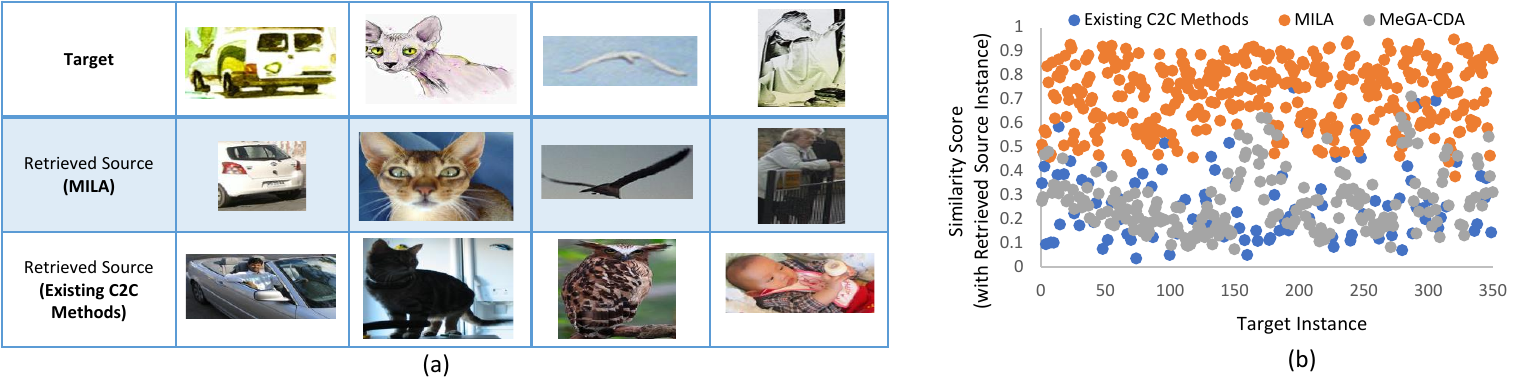}
      \caption{(a) Examples of retrieved source instances by MILA and the C2C method for alignment. Our approach retrieves source instances with similar visual characteristics compared to existing C2C methods. (b) Comparison of similarity scores among selected cross-domain pairs for alignment using different methods reveals that MILA aligns a target instance only with a source instance that has a very high similarity score.} 
  \label{fig:fig1}
\end{figure*}

Even though C2C methods achieve better performance than vanilla instance alignment methods, we argue that aligning a target instance in a category to an arbitrary source instance in the same category is sub-optimal. 
We claim that it is important to align a target instance in a category to the source instance in the same category that is similar in terms of the {\it non-defining} visual characteristics.
We assume that the characteristics of an image in the view of the object detection task are divided into three groups: {\it defining characteristics}, which are indispensable for defining a category (\eg, shape), {\it non-defining characteristics}, which can be different and diverse within a category (\eg, orientation, color), and {\it domain specific characteristics}, which can be different within a category but shared within a domain (\eg, style).
In the domain adaptation process, it is important to focus on aligning domain specific features and aligning non-defining features are not necessary.
By finding a source instance that has similar non-defining characteristics with a target instance, a model is not disturbed by the unimportant differences and can focus solely on the difference in domains. 
The existing C2C methods often struggle to find a suitable source instance because they search a matching instance only within a mini-batch, which does not necessarily contain a suitable source instance since mini-batches are usually small in size and thus tend to lack diversity of the samples (see \Cref{fig:fig1}).

To address this issue, we propose memory-based instance-level adaptation (MILA).
We design MILA in such a way that it can learn alignment from the {\it `reliable'} matching pairs.
A pair of source and target instance is regarded as {\it reliable} when {\it (i)} their features are expected to well represent the categories and {\it (ii)} these features are similar enough so that a model can focus on domain differences.
To increase the chance of finding reliable matching pairs, MILA has four unique characteristics. 
(1) MILA has a memory module for storing source features, which is much larger than mini-batch and thus greatly increases chance to find a suitable source instance for the alignment (contribution to {\it (ii)}).
(2) MILA stores source features only when the model can correctly predict the category of the source instance using the features.
This is to guarantee the quality of the stored features (contribution to {\it (i)}).
(3) MILA uses a target instance for the alignment only when the category prediction confidence is sufficiently high.
This is to guarantee the quality of the target features (contribution to {\it (i)}).
(4) MILA assigns different weights to different matching pairs according to their similarities to emphasize more reliable pairs (contribution to {\it (ii)}).

Our main contributions are summarized into 3 points. 
(1) We are the first, to the best of our knowledge, to argue the importance of finding `reliable' pairs for the domain-adaptive object detection (DAOD) task, and provide with the empirical evidence for it (\Cref{fig:fig1}).
(2) We propose a dedicated design based on the memory module for increasing the chance of finding `reliable' pairs.
(3) We verify the effectiveness of the proposed method and its four characteristics by extensive experiments.
It achieved state-of-the-art results on five cross-domain object detection tasks, with significant relative improvements of 5.5\%, 4.1\%, and 4.0\% on the \textit{Sim10k}, \textit{Comic2k} and \textit{Watercolor2k} benchmark datasets, respectively.

\section{Related Works}
\label{sec:related}
\paragraph{Domain Adaptive Object Detection (DAOD).} DA-Faster~\cite{chen2018domain} proposed an early DAOD method that performs feature alignment at both
image-level and instance-level. MAF~\cite{he2019multi} and~\cite{xie2019multi} extended this idea to multi-layer feature adaptation of backbone network. SWDA~\cite{saito2019strong} suggests that aligning local features strongly is more effective than aligning global features strongly. CRDA~\cite{xu2020exploring} and MCAR~\cite{zhao2020adaptive} introduce a multi-label classifier upon backbone network to regularize the features. Recent methods~\cite{xu2020cross, he2020domain, li2020spatial, zhang2021rpn, vs2021mega, su2020adapting, zhu2019adapting} align instance-level features from object proposals using category-aware manner (C2C). They derive prototype representation of each category by aggregating multiple instances before inducing alignment. However, this causes loss of intra-class variance information and leads to suboptimal prototypes for alignment.
%
%
%
\paragraph{Memory Networks.}
The memory network~\cite{weston2014memory, sukhbaatar2015end} is a type of neural network module that utilizes an external memory to store and retrieve relevant information. It has been widely utilized in vision-related tasks, such as video object segmentation~\cite{rodriguez2019domain, oh2019video}, movie understanding~\cite{na2017read}, and visual tracking~\cite{yang2018learning} due to its ability to retain diverse knowledge types.
%
%
%

Memory networks have also been used in domain adaptation~\cite{memsac}, and DAOD~\cite{vs2021mega}. The closest to our work is MeGA-CDA~\cite{vs2021mega}, which utilizes memory modules for storing class-prototypes and use them to create category-specific attention maps for better C2C alignment between source and target instances.
Although MeGA-CDA is similar to MILA in a sense that both use memory module, their motivations differ significantly. 
MeGA-CDA is the method dedicated for C2C alignment and it aims to find source regions that correspond to a particular category within a limited search scope of each mini-batch, while MILA is designed to find most `reliable' source instance of a category for the alignment and exploit more from reliable pairs.
In other words, MeGA-CDA cares only categories, while MILA also cares specific instances in addition to categories.
MILA has many unique characteristics to increase the chance of finding reliable matching pairs, as mentioned in the introduction.
In fact, sometimes MeGA-CDA fails to find the appropriate source regions to match with a given target sample because there is no guarantee that a mini-batch always contains an object of a particular category.
In contrast, MILA can store wide variety of source instances of all the categories in the memory and therefore can always find a source instance of the same category as a given target instance.



\begin{figure*}[t!]
  \centering
  \includegraphics[width=\textwidth]{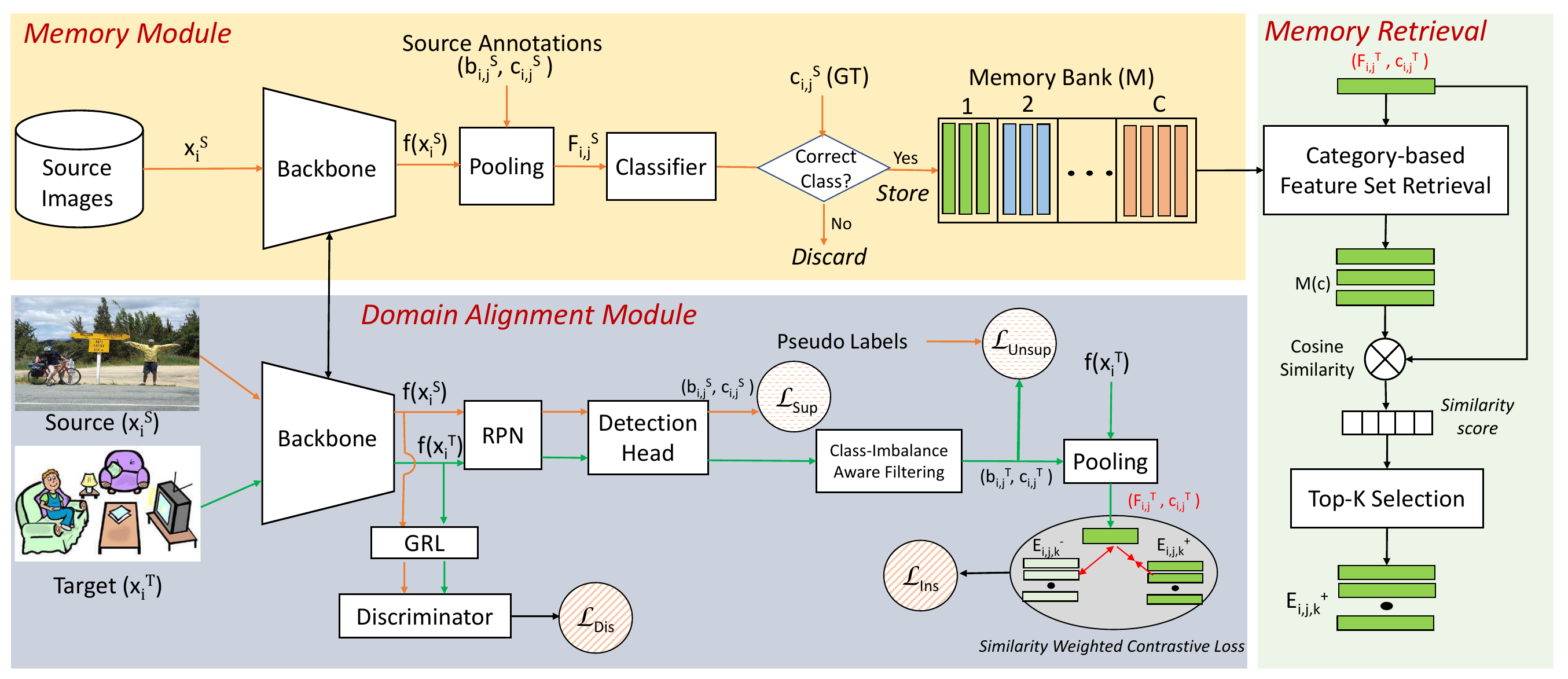}
  \caption{Network Overview: Mainly consist of memory module,  similarity-based memory retrieval module, and instance-level domain adaptation module}
  \label{fig:fig2}
\end{figure*}

\section{Method}


\subsection{ Problem Formulation}\label{sec:problem}
Given a labeled source dataset $D_S = \{x^S_i, y^S_i\}_{i=1}^{N_S}$ and an unlabeled target dataset $D_T = \{x^T_i\}_{i=1}^{N_T}$, the task of UDA is to transfer knowledge from $D_S$ to $D_T$ and predict accurate labels for $D_T$. Even though $D_S$ and $D_T$ have the same label space, they come from very different data distributions, which makes UDA a very challenging task. Since we focus on UDA for object detection, $y^S_{i} = \{b^S_{i,j}, c^S_{i,j}\}_{j=1}^{K^S_i}$ contains bounding box labels $b^S_{i,j}\in [0, 1]^4$ and category labels $c^S_{i,j} \in \{1, \dots, C\}$ for $K^S_i$ objects in image $x^S_i$, where there are $C$ different categories of objects.

\subsection{Overview}\label{sec:overview}

To overcome the limitations of existing C2C alignment methods discussed in \cref{sec:intro}, we present MILA. \Cref{fig:fig2} illustrates our proposed approach, which is built on Faster R-CNN, following prior works~\cite{saito2019strong, xu2020cross, li2022cross}. MILA includes three main modules: (1) a \textit{memory module} that stores instance-level features and category information from previously-seen source images, (2) a \textit{memory retrieval module} that retrieve most `reliable' source instances of the same category as a given target instance feature,
and (3) a \textit{similarity-weighted alignment} that scales target instance alignment based on similarity to retrieved source instances. Detailed descriptions of these modules are provided in the following sections.
%
We consider the two-stage Faster-RCNN to be made up of a feature extractor $f( ; \theta)$ followed by a region-proposal network $rpn(; \phi)$ and a detection head $det(;\psi)$, where $\theta, \phi$ and $\psi$ are learnable parameters.

\subsection{Memory Module}\label{sec:memory module}
The memory module is used to save the instance-level features of source images extracted from the Faster R-CNN backbone in a memory bank. Specifically, for a source image $x^S_i$, $f$ computes the feature map as $f(x^S_i; \theta)$. 
The features for $j^{th}$ instance in the image can be extracted from $f(x^S_i; \theta)$ by pooling corresponding instance bounding box $b^S_{i,j}$ available in $y^S_i$. Therefore,
\begin{equation}
    \hat{F}^S_{i,j}= roipool\left(f(x^S_i; \theta), b^S_{i,j}\right) \hspace{1mm},
\end{equation}
where $roipool$ is the conventional region-of-interest (ROI) pooling function used in Faster R-CNN. The extracted instance features $\hat{F}^S_{i,j}$ is then filtered based on the classification accuracy as follows:

\begin{equation}
    F^S_{i,j} = \begin{cases} \hat{F}^S_{i,j} & \text{if } \hat{c}^S_{i,j} = c^S_{i,j} \\
discard & \text{otherwise}
\end{cases}
\end{equation}
where $\hat{c}^S_{i,j}$ is the predicted class of the source proposal $\hat{F}^S_{i,j}$ of original class label $c^S_{i,j}$.
%
The filtered instance features $F^S_{i,j}$ are category-wise stored in the memory bank, which is created as follows:
\begin{equation}
    M(c) = \left\{\left\{\mathbf{I}\left(c^S_{i,j} = c\right)F^S_{i,j}\right \}_{j=1}^{K^S_i}\right\}_{i=1}^{N_S} \hspace{1mm} ,
\end{equation}
where $\mathbf{I}$ denotes the indicator function and $M(c)$ denotes all the stored instance features of category $c$, where $1 \leq c \leq C$.



During training, as $\theta$ gets updated and $f$ extracts more domain-aligned features, we propose dynamically updating the memory to store high-quality representations. Note that the memory $M$ is only created during training.

\subsection{Similarity-Based Memory Retrieval Module} \label{sec:retrieval module}
%
This module's objective is to retrieve `reliable' source instances that have similar features to the target instance, allowing the model to focus on domain differences.
The retrieval process is structured as follows: (1) Predicting the bounding boxes and their categories in the target image, and filtering out the inaccurate predictions. (2) Extracting the instance-level features from the predicted bounding boxes in the target images. (3) Retrieving the top-$K$ similar source instance features from memory $M$ that correspond to each target instance feature.
%



 
For $i^{th}$ target image $x^T_i$, the feature map is generated as $f(x^T_i; \theta)$. This feature map is then used by $rpn(;\phi)$ and $det(;\psi)$ to predict $\hat{y}^T_{i}$, which has $K^T_i$ instances with bounding box, category, and confidence score, i.e., 
\begin{equation}
    \hat{y}^T_{i} =\{b_{i,j}^T, c_{i,j}^T, s_{i,j}^T\}_{j=1}^{K^T_i} = det( rpn( f(x^T_i; \theta); \phi); \psi)
\end{equation}

To ensure MILA's effectiveness, it's crucial to filter out noisy predictions as they can hamper performance. Existing methods~\cite{li2022cross} use a fixed threshold $\delta$ for all classes to remove predictions with confidence scores ($s_{i,j}^T$) below $\delta$. However, due to class imbalance in the training data, lower confidence scores are often produced for underrepresented classes, making a fixed threshold impractical. 
To address this challenge, we assign category-specific thresholds $\delta_{c}$ based on the detection accuracy of each category in the source dataset inspired by \cite{sinha2020class}.
Note before filtering, we perform non-maximum suppression (NMS) $\sigma$ to remove duplicate bounding boxes, i.e.,

\begin{equation}
    \tilde y^T_{i} =  \left\{\sigma \left(b_{i,j}^T, c_{i,j}^T, s_{i,j}^T\right) \middle| s^T_{i,j} \geq \delta_{c}\right\}^{j=K^T_i}_{j=1}   \hspace{1mm},
\end{equation}
Once we have the filtered bounding boxes in $\tilde y^T_{i}$, we can extract instance-level features from $f(x^T_i; \theta)$ by pooling from the predicted bounding boxes $b^T_{i,j}$ ($j^{th}$ box for $i^{th}$ target image) as
\begin{equation}
    F^T_{i,j} = roipool(f(x^T_i; \theta), b^T_{i,j}) \hspace{1mm},
\end{equation}

From memory $M$, we select the source features to be used for the alignment with $F^T_{i,j}$ from the features that has the same category as the predicted category $c_{i,j}^T$.
The cosine similarity scores of $F^T_{i,j}$ with each source feature of the same category is computed as
\begin{equation}
    S(F^T_{i,j}, M(c^T_{i,j})_z) = \frac{F^T_{i,j} . M(c^T_{i,j})_z}{\parallel F^T_{i,j}\parallel \parallel M(c^T_{i,j})_z\parallel} \hspace{1mm},
\end{equation}
where $M(c)_z$ is the $z^{th}$ source instance feature stored in $M(c)$.
Now, we can easily retrieve top-$K$ most similar source instance features of the same category as the target instance $c^T_{i,j}$.
We call them positive samples and denote by $E^{+}_{i,j,k}$, where $1\leq k \leq K$ represents the index. Similarly, negative samples $E^{-}_{i,j,k}$ are obtained by randomly selecting 
one sample from each of the categories that are different from the category of the positive pairs. 

\subsection{Similarity-weighted Instance-level Domain Adaptation} \label{sec:alignment}
Once we retrieve positive and negative set of instances from source memory for a given target instance, we align them by applying a specially designed max-margin contrastive losses. 
For a target instance feature $F_{i,j}^{T}$ and $k$-th positive sample $E^{+}_{i,j,k}$, contrastive loss enforces them to come closer in latent space, whereas pushes the features apart for negative samples $E^{-}_{i,j,k}$.

\begin{equation}
\begin{gathered}
\mathcal{L}^{+}_{i,j} =\frac{1}{K}\sum_{k=1}^{K}S(F^T_{i,j}, E^+_{i,j,k})d_{pos}, \hspace{10pt}
\mathcal{L}^{-}_{i,j} = \frac{1}{C-1}\sum_{k=1}^{C-1}max(0, m - d_{neg}),\\
\mathcal{L}_{Ins} = \frac{1}{N^T}\sum_{i=1}^{N^T} \left( \frac{1}{K^T_i} \sum_{j=1}^{K^T_i}\mathcal{L}^{+}_{i,j} + \mathcal{L}^{-}_{i,j} \right). \\
\label{eqn:contrastive loss}
\end{gathered}
\end{equation}
Where $d_{pos}$ and $d_{neg}$ are the Euclidean distances of $F_{i,j}^{T}$ with $E^{+}_{i,j,k}$ and $E^{-}_{i,j,k}$, respectively. Recall that cardinality of the negative sample sets is $C-1$, i.e., $|E^{-}_{i,j,k}|=C-1$.

\subsection{Overall Objective}\label{sec:overall loss}
In addition to the instance-level alignment loss discussed earlier, we use supervised loss $\mathcal{L}_{Sup}$, unsupervised loss $\mathcal{L}_{Unsup}$, and discriminator loss $\mathcal{L}_{Dis}$.
$\mathcal{L}_{Sup}$ is the object detection loss optimized using labeled data from the source domain.
$\mathcal{L}_{Unsup}$ 
is the object detection loss computed on target domain images with pseudo labels generated using a similar approach as in~\cite{li2022cross}.
$\mathcal{L}_{Dis}$ is the loss of an image-level binary domain discriminator used in~\cite{he2019multi, xie2019multi}.
$\mathcal{L}_{Dis}$ is computed by a domain discriminator $D$ whose aim is to discriminate where the backbone feature is from ($d = 0$) or target ($d = 1$). 
%
$\mathcal{L}_{Dis}$ is formulated as follows:
\begin{align*}
  \MoveEqLeft[5] \mathcal{L}_{Dis} = \frac{1}{(N^T+N^S)} \sum_{i=1}^{N^T+N^S}- d\log D(f(x_{i}; \theta))-(1-d)\log(1-D(f(x_{i}; \theta))),
\end{align*}
Note that the gradients obtained from this loss term are used to update not only the parameters of the discriminator $D$, but also reversed by the gradient reversal layer (GRL) during backpropagation to update $\theta$. This helps $f(;\theta)$ to learn discriminative features that can confuse $D$.
Combining altogether, the overall objective function is as follows:
\begin{equation}
\mathcal{L}=\mathcal{L}_{Sup} + \lambda_1 \mathcal{L}_{Unsup}+ \lambda_2 \mathcal{L}_{Dis} + \lambda_3 \mathcal{L}_{Ins},
\label{eq:overall_loss}
\end{equation}
where $\lambda_1$, $\lambda_2$ and $\lambda_3$ are the hyperparameters to control the weight of each loss.

\begin{table}
\begin{minipage}[t]{.45\textwidth}
  \footnotesize	
  \centering
  \addtolength{\tabcolsep}{-3.0pt}
  \scalebox{0.7}{
    \begin{tabular}{l|cccccc|r}
      \toprule 
      Method & bicycle & bird & car & cat & dog & person & mAP\\
      \midrule 
      Source & 32.5 & 12.0 & 21.1 & 10.4 & 12.4 & 29.9 & 19.7 \\
       \midrule 
      DA-Faster~\cite{chen2018domain} & 31.1 & 10.3 & 15.5 & 12.4 & 19.3 & 39.0 & 21.2 \\
      SWADA~\cite{saito2019strong} & 36.4 & 21.8 & 29.8 & 15.1 & 23.5 & 49.6 & 29.4 \\
      MCAR~\cite{zhao2020adaptive} & 49.7 & 20.5 & 37.4 & 20.6 & 24.5 & 53.6 & 33.5 \\
      D-Adapt~\cite{jiang2021decoupled} & 52.4 & 25.4 & 42.3 & \textbf{43.7} & 25.7 & 53.5 & 40.5 \\
      \midrule 
      Ours & \textbf{59.1} & \textbf{28.5} & \textbf{49.8} & 28.3 & \textbf{35.7} & \textbf{66.3} & \textbf{44.6}\\
      \midrule 
      Oracle & 44.2 & 35.3 & 31.9 & 46.2 & 40.9 & 70.9 & 44.6\\
      
      \bottomrule 
    \end{tabular}
    }
  \caption{Results on the Comic2k test set for \textbf{Pascal VOC$\xrightarrow{}$Comic2k} adaptation (ResNet-101).} 
   \label{tab:table1_}
\end{minipage}\qquad
\begin{minipage}[t]{.45\textwidth}
  \footnotesize	
  \addtolength{\tabcolsep}{-3.0pt}
  \scalebox{0.72}{
    \begin{tabular}{l|cccccc|r}
      \toprule 
      Method & bicycle & bird & car & cat & dog & person & mAP\\
      \midrule 
      Source & 84.2 & 44.5 & 53.0 & 24.9 & 18.8 & 56.3 & 46.9\\ 
       \midrule 
      SCL~\cite{shen2019scl} & 82.2 & 55.1 & 51.8 & 39.6 & 38.4 & 64.0 & 55.2 \\
      SWADA~\cite{saito2019strong} & 82.3 & 55.9 & 46.5 & 32.7 & 35.5 & 66.7 & 53.3 \\
      UMT~\cite{deng2021unbiased} & 88.2 & 55.3 & 51.7 & 39.8 & 43.6 & 69.9 & 58.1 \\
      AT~\cite{li2022cross} & 94.3 & 57.2 & 57.2 & 34.2 & 36.9 & 78.5 & 59.7 \\ 
      \midrule 
      Ours & \textbf{97.4} & \textbf{59.0} & \textbf{58.3} & \textbf{40.6} & \textbf{47.8} & \textbf{79.3} & \textbf{63.7}\\ 
      \midrule 
      Oracle & 51.8 & 49.7 & 42.5 & 38.7 & 52.1 & 68.6 & 50.6\\
      
      \bottomrule 
    \end{tabular}
   }
  \caption{Results on the Watercolor2k test set for \textbf{Pascal VOC$\xrightarrow{}$Watercolor2k} adaptation (ResNet-101).} 
   \label{tab:table3}
\end{minipage}
\end{table}

\section{Experiments}
\subsection{Datasets}
We performed extensive experiments on seven publicly available datasets covering five scenarios of domain shift. The datasets are Pascal VOC~\cite{everingham2010Pascal}, Clipart1k~\cite{inoue2018cross}, 
Watercolor2k~\cite{inoue2018cross}, 
Comic2k~\cite{inoue2018cross}, 
Sim10k~\cite{johnson2016driving}, Cityscapes~\cite{cordts2016cityscapes}, and FoggyCityscapes~\cite{sakaridis2018semantic}. Pascal VOC comprises 16,551 images of 20 categories of common objects from the real world. Clipart1k contains 1k comical images with the same 20 categories as Pascal VOC. Watercolor2k includes 1k training and 1k testing watercolor-style images, sharing six categories with Pascal VOC. Similarly, Comic2k consists of 1k training and 1k test images, sharing six categories with Pascal VOC. Sim10k contains 10,000 images with 58,701 bounding boxes of car categories, while both Cityscapes and FoggyCityscapes comprise 2,975 training images and 500 validation images with eight object categories. We evaluate the domain adaptation performance of different methods using the standard setting~\cite{saito2019strong, li2022cross} on the following five domain adaptation tasks: (i) Pascal VOC$\xrightarrow{}$Comic2k, (ii) Pascal VOC$\xrightarrow{}$Watercolor2k, (iii) Pascal VOC$\xrightarrow{}$Clipart1k, (iv) Sim10k$\xrightarrow{}$Cityscapes, and (v) Cityscapes$\xrightarrow{}$FoggyCityscapes.
Due to space constraints, we provide results for Pascal VOC$\xrightarrow{}$Clipart1k in supplementary materials.

\subsection{Implementation Details}\label{sec:implementation_details}
We adopt the Faster R-CNN model with either ResNet-101 or VGG16 architectures and implement it using Detectron2, following the approach in~\cite{saito2019strong, xu2020cross, tian2021knowledge, li2022cross}. We fine-tune the parameters of ResNet-101 from the model pre-trained on ImageNet~\cite{deng2009imagenet}, while VGG16 parameters are learned from scratch. The images are scaled by resizing the shorter side to 600 pixels while maintaining the aspect ratios, following the common practice~\cite{he2019multi}. We apply a set of strong and weak data augmentations as described in~\cite{li2022cross}. Our evaluation metric is the average precision (AP) for each class and the mean AP (mAP) over all classes.
For the hyperparameters, we set $\lambda_1$ to $1.0$, $\lambda_2$ to $0.1$, and $\lambda_3$ to $0.1$ unless otherwise stated. Margin $m$ in~\cref{eqn:contrastive loss} is set to $1.0$. We update the memory after every $1/3$ of an epoch and use stochastic gradient descent (SGD) with momentum 0.9 and a learning rate of 0.04 throughout the training stage, without any learning rate decay. We build on the code provided by the authors of~\cite{li2022cross} and follow their hyperparameter settings. The experiments are conducted on 4 Nvidia GPU V100 with a batch size of 8 or 4 depending on the dataset, using PyTorch.

\subsection{Comparison with state-of-the-arts}
We compare the proposed MILA with the state-of-the-art DAOD methods, including SCL~\cite{shen2019scl}, SWADA~\cite{saito2019strong}, DM~\cite{kim2019diversify}, CRDA~\cite{xu2020exploring}, HTCN~\cite{chen2020harmonizing}, DA-Faster~\cite{chen2018domain}, MCAR~\cite{zhao2020adaptive}, D-Adapt~\cite{jiang2021decoupled},
MAF~\cite{he2019multi}, SCDA~\cite{zhu2019adapting}, CDN~\cite{li2020spatial}, MeGA-CDA~\cite{vs2021mega}, CADA~\cite{hsu2020every}, UMT~\cite{deng2021unbiased}, and Adaptive Teacher (AT)~\cite{li2022cross}.
%
%
%
Our implementation, as described in Sec.~\ref{sec:implementation_details}, builds upon the official code of \cite{li2022cross}. To ensure fairness, we compared our method with the best-reproduced results of AT, under the same conditions. The original results from~\cite{li2022cross} are in the supplementary material. `Source' is the baseline model without domain adaptation, while `Oracle' is trained and tested on the target domain.  

\begin{table}
\begin{minipage}[t]{.5\textwidth}
\centering 
  \footnotesize	
  \addtolength{\tabcolsep}{-3.5pt}
  \scalebox{0.6}{
  \begin{tabular}{l|cccccccc|r}
      \toprule 
      Method & bus & bicycle & car & mcycle & person & rider & train & truck & mAP\\
      \midrule 
      Source (F-RCNN) & 20.1 & 31.9 & 39.6 & 16.9 & 29.0 & 37.2 & 5.2 & 8.1 & 23.5 \\
       \midrule 
      SCL~\cite{shen2019scl} & 41.8 & 36.2 & 44.8 & 33.6 & 31.6 & 44.0 & 40.7 & 30.4 & 37.9 \\
      DA-Faster~\cite{chen2018domain} & 35.3 & 27.1 & 40.5 & 20.0 & 25.0 & 31.0 & 20.2 & 22.1 & 27.6 \\
      SCDA~\cite{zhu2019adapting} & 39.0 & 33.6 & 48.5 & 28.0 & 33.5 & 38.0 & 23.3 & 26.5 & 33.8 \\
      SWDA~\cite{saito2019strong} & 36.2 & 35.3 & 43.5 & 30.0 & 29.9 & 42.3 & 32.6 & 24.5 & 34.3 \\
      DM~\cite{kim2019diversify} & 38.4 & 32.2 & 44.3 & 28.4 & 30.8 & 40.5 & 34.5 & 27.2 & 34.6 \\
      MTOR~\cite{cai2019exploring} & 38.6 & 35.6 & 44.0 & 28.3 & 30.6 & 41.4 & 40.6 & 21.9 & 35.1 \\
      MAF~\cite{he2019multi} & 39.9 &  33.9 & 43.9 & 29.2 & 28.2 & 39.5 & 33.3 & 23.8 & 34.0 \\
      iFAN~\cite{zhuang2020ifan} & 45.5 & 33.0 & 48.5 & 22.8 & 32.6 & 40.0 & 31.7 & 27.9 & 35.3 \\
      CRDA~\cite{xu2020exploring} & 45.1 & 34.6 & 49.2 & 30.3 & 32.9 & 43.8 & 36.4 & 27.2 & 37.4 \\
      HTCN~\cite{chen2020harmonizing} & 47.4 & 37.1 & 47.9 & 32.3 & 33.2 & 47.5 & 40.9 & 31.6 & 39.8 \\
      UMT~\cite{deng2021unbiased} & 56.5 & 37.3 & 48.6 & 30.4 & 33.0 & 46.7 & 46.8 & 34.1 & 41.7 \\
      AT~\cite{li2022cross} & 60.0 & 49.0 & 63.6 & 38.8 & 45.0 & \textbf{53.9} & 45.1 & 33.9 & 49.0 \\
      \midrule
      Ours & \textbf{61.4}  & \textbf{51.5} & \textbf{64.8} & \textbf{39.7} & \textbf{45.6} & \textbf{52.8} & \textbf{54.1} & \textbf{34.7} & \textbf{50.6} \\
      \midrule 
       Oracle (F-RCNN) & 50.3 & 40.7 & 61.3 & 32.5 & 43.1 & 49.8 & 35.1 & 28.6 & 42.7\\
      \bottomrule 
    \end{tabular}
    }
    \caption{Results on the FoggyCityscapes test set for \textbf{Cityscapes $\xrightarrow{}$ Foggy Cityscapes} adaptation (VGG-16).} 
   
   \label{tab:table2}
\end{minipage}\qquad
\begin{minipage}[t]{.45\textwidth}
  \footnotesize	
  \centering
  \addtolength{\tabcolsep}{-3.0pt}
  \scalebox{0.68}{
   \begin{tabular}{l|c|r}
      \toprule 
      Method & Backbone & AP on Car\\
      \midrule 
      DA-Faster~\cite{chen2018domain} &  & 38.9 \\
      BDC-Faster~\cite{saito2019strong} &  & 31.8 \\
      SWADA~\cite{saito2019strong} &  & 40.1 \\
      MAF~\cite{he2019multi} &   & 41.1 \\
      SCDA~\cite{zhu2019adapting} &   & 43.0 \\
      CDN~\cite{li2020spatial} &  VGG-16 & 49.3 \\
      MeGA-CDA~\cite{vs2021mega} &   & 44.8 \\
      UMT~\cite{deng2021unbiased} &   & 43.1\\
      \midrule 

      Source &   & 41.8 \\
      CADA~\cite{hsu2020every} &  & 51.2 \\
      D-adapt~\cite{jiang2021decoupled} & ResNet-101 & 51.9\\
      Ours &  & \textbf{57.4} \\
      \midrule 
      Oracle &  & 70.4\\
      \bottomrule 
    \end{tabular}
    }
   \caption{Results on the Cityscapes test set for \textbf{Sim10k $\xrightarrow{}$ Cityscapes} adaptation.} 
   \label{tab:table2_}
\end{minipage}\qquad
\end{table}

\paragraph{Adaptation between dissimilar domains.}
%
%
In our first set of experiments, we use the Pascal VOC as the source domain and Comic2k as the target domain, which has a very different style from Pascal VOC and contains many small objects. Table~\ref{tab:table1_} shows that MILA outperforms all the previous methods and improves mAP by 4.1 compared with the state-of-the-art. 
%
%
\\[2.5pt]
In the next set of experiments, we evaluate MILA on Watercolor2k, another target domain with a unique style. As shown in Table~\ref{tab:table3}, MILA achieves the highest average precision for all object categories, surpassing all previous methods by a significant margin. Additionally, MILA surpasses the oracle model on the Watercolor2k dataset by a considerable margin of $13.1\%$. 
%
%
These results consistently validate the effectiveness of aligning the most similar instances across domains in reducing the domain gap between different scenarios.

\paragraph{Adaptation between similar domains.}
The results of this setting are presented in Table~\ref{tab:table2}. MILA achieves the highest mAP in majority of the categories. Upon closer inspection, we observed that MILA achieved the largest performance gain of $+9.0\%$ in `train' class, which has the least number of instances, with only 504 training samples. This finding suggests that classes with fewer training instances specially benefit from the proposed memory module. We hypothesize that this is because it is generally more challenging to align less populated classes due to the difficulty of finding an appropriate alignment target. MILA helps to overcome this challenge by storing all the alignment targets in the memory.

\paragraph{Adaptation from synthetic to real images.}
We use Sim10k as the source domain and Cityscapes as the target domain. Following~\cite{jiang2021decoupled}, we evaluate on the validation split of the Cityscapes and report the mAP on car. Table~\ref{tab:table2_} shows that MILA scores the new state-of-the-art adaptation performance, achieving gain of 5.5 points on mAP.

\begin{figure*}[t!]
  \centering
  \includegraphics[width=\textwidth]{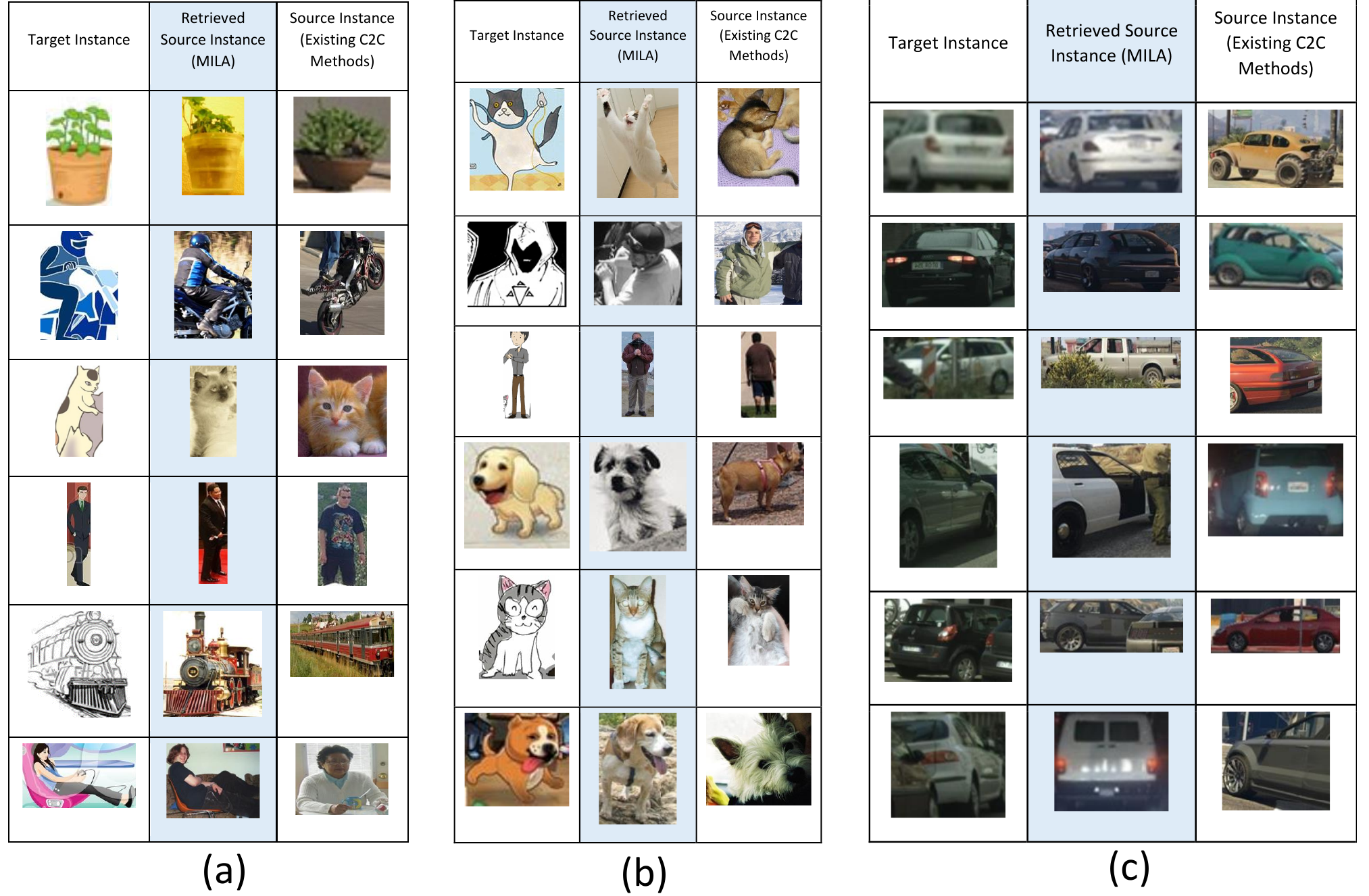}
  \caption{Visualization of instance pairs (a) Pascal VOC$\xrightarrow{}$Clipart1k (b) Pascal VOC$\xrightarrow{}$Comic2k (c) Sim10k$\xrightarrow{}$Cityscapes}
  \label{fig:fig100}
\end{figure*}

\subsection{Visualization of Instance Pairs}\label{sec:qualitative}
To visualize reliable matching pairs used in alignment, we present target instances retrieved by MILA in \cref{fig:fig100}.
In the second example of \cref{fig:fig100} (a), MILA retrieves a biker instance with a matching color scheme for the helmet and bike as the target instance. Similarly, in the fourth example (\cref{fig:fig100} (a)), MILA successfully identifies a person wearing a matching dress, showcasing its ability to capture subtle visual details essential for effective domain adaptation. 

On other datasets, MILA consistently demonstrates its capability to retrieve instances with remarkable visual similarities. For example, in \cref{fig:fig100} (b), the first and fifth example demonstrate MILA's ability to retrieve instances of the cat category that not only share similar color and orientation but also exhibit an overall appearance that closely matches the target instances.
In \cref{fig:fig100} (c), MILA retrieve car instances with matching color and orientation, as evident in examples 1, 2, and 6 for rear-facing target cars and examples 3, 4, and 5 for side-facing cars. In contrast, existing C2C methods retrieve source instances of car that display significant differences in color and orientation when compared to the target instances.

These visual examples highlight MILA's exceptional capability in identifying source instances with similar non-defining visual characteristics as the target compared to existing C2C methods. By focusing solely on domain differences and disregarding unimportant dissimilarities, our model achieves superior accuracy.

\begin{wraptable}{l}{0.45\textwidth}
\footnotesize
\centering
\addtolength{\tabcolsep}{-4.0pt}
 \scalebox{0.9}{
\begin{tabular*}{0.45 \textwidth}{@{\extracolsep{\fill}\quad}l|ccccc}
\toprule
mAP & $m$ & $f$ & $s$ & $t$ & $u$ \\
\midrule
$44.6$ & \checkmark & \checkmark & \checkmark & \checkmark & \xmark \\
$40.3$ &\xmark & \xmark & \xmark & \xmark & \checkmark  \\
$42.6$ &\checkmark & \xmark & \checkmark & \checkmark & \xmark \\
$42.2$ & \checkmark & \checkmark & \xmark & \checkmark & \xmark   \\
$43.9$ &\checkmark & \checkmark & \checkmark & \xmark & \checkmark   \\

\bottomrule
\end{tabular*}
}
 \caption{Ablation study of different components. } 
    \label{tab:tableab_}
\end{wraptable}

\subsection{Ablation Study}\label{sec:ablation}
In this section, we assess effectiveness of different components in our approach on the Pascal VOC $\rightarrow$ Comic2k dataset (see \cref{tab:tableab_}). We used our full model with all four characteristics discussed in ~\cref{sec:intro} and turned off each component one by one to evaluate their impact on the detection accuracy of the model.

First, we evaluate the \textbf{memory module's effectiveness ($m$)} by comparing the performance of our model with and without the memory module. We used a non-memory based instance alignment scheme~\cite{xu2020cross} instead and used the same contrastive loss function $\mathcal{L}_{Ins}$, but sampled positive and negative instances from the mini-batch instead of the memory module. Our results showed that MILA improves object detection accuracy by 4.3\% compared to the model without the memory module ($2^{nd}$ row), indicating the importance of the memory module in enhancing performance. 
\\
Secondly, we analyze the \textbf{effectiveness of source feature filtering ($f$)} to ensure the quality of the stored features in memory by checking their predicted category. We observed that the performance drops by $2.0$ points ($3^{rd}$ row) if we turn off this component and store all source features in memory without judging their quality.
\\
Thirdly, we evaluate the \textbf{effectiveness of similarity weighting of contrastive loss ($s$)} ($4^{th}$ row). We observe that if we use plain contrastive loss function, there is a drop in accuracy ($2.2\%$), which suggests that weighting our loss function by similarity helps to mitigate negative knowledge transfer when the aligned features are highly dissimilar. As a result, the detection accuracy improves.
\\
We also analyzed the \textbf{effectiveness of category-aware thresholding ($t$) and fixed thresholding ($u$)}. The model performed slightly better with category-imbalance aware thresholding, compared to the fixed thresholding. 
\\
In summary, our results demonstrate that all four components are critical for our approach, and turning them off results in decreased detection accuracy. Further analysis on the memory module and hyperparameters are available in the supplementary material.


\section{Conclusion}
In this paper, we propose a Memory-based Instance-Level Alignment (MILA) framework for cross-domain object detection. 
The proposed strategy with four unique characteristics enables the model in finding ‘reliable’ pairs for alignment in the domain-adaptive object detection (DAOD) task. As evident in the results, this can efficiently improve the adaptation of instances by only focusing on important visual characteristics that distinguish the two domains. Extensive experiments demonstrate that MILA achieves state-of-the-art performance for adapting object detectors on several benchmark datasets.


\bibliography{egbib}

\begin{thebibliography}{40}
\providecommand{\natexlab}[1]{#1}
\providecommand{\url}[1]{\texttt{#1}}
\expandafter\ifx\csname urlstyle\endcsname\relax
  \providecommand{\doi}[1]{doi: #1}\else
  \providecommand{\doi}{doi: \begingroup \urlstyle{rm}\Url}\fi

\bibitem[Cai et~al.(2019)Cai, Pan, Ngo, Tian, Duan, and Yao]{cai2019exploring}
Qi~Cai, Yingwei Pan, Chong-Wah Ngo, Xinmei Tian, Lingyu Duan, and Ting Yao.
\newblock Exploring object relation in mean teacher for cross-domain detection.
\newblock In \emph{CVPR}, 2019.

\bibitem[Chen et~al.(2020)Chen, Zheng, Ding, Huang, and
  Dou]{chen2020harmonizing}
Chaoqi Chen, Zebiao Zheng, Xinghao Ding, Yue Huang, and Qi~Dou.
\newblock Harmonizing transferability and discriminability for adapting object
  detectors.
\newblock In \emph{CVPR}, 2020.

\bibitem[Chen et~al.(2018)Chen, Li, Sakaridis, Dai, and
  Van~Gool]{chen2018domain}
Yuhua Chen, Wen Li, Christos Sakaridis, Dengxin Dai, and Luc Van~Gool.
\newblock Domain adaptive faster r-cnn for object detection in the wild.
\newblock In \emph{CVPR}, 2018.

\bibitem[Cordts et~al.(2016)Cordts, Omran, Ramos, Rehfeld, Enzweiler, Benenson,
  Franke, Roth, and Schiele]{cordts2016cityscapes}
Marius Cordts, Mohamed Omran, Sebastian Ramos, Timo Rehfeld, Markus Enzweiler,
  Rodrigo Benenson, Uwe Franke, Stefan Roth, and Bernt Schiele.
\newblock The cityscapes dataset for semantic urban scene understanding.
\newblock In \emph{CVPR}, 2016.

\bibitem[Deng et~al.(2009)Deng, Dong, Socher, Li, Li, and
  Fei-Fei]{deng2009imagenet}
Jia Deng, Wei Dong, Richard Socher, Li-Jia Li, Kai Li, and Li~Fei-Fei.
\newblock Imagenet: A large-scale hierarchical image database.
\newblock In \emph{CVPR}, 2009.

\bibitem[Deng et~al.(2021)Deng, Li, Chen, and Duan]{deng2021unbiased}
Jinhong Deng, Wen Li, Yuhua Chen, and Lixin Duan.
\newblock Unbiased mean teacher for cross-domain object detection.
\newblock In \emph{CVPR}, 2021.

\bibitem[Everingham et~al.(2010)Everingham, Van~Gool, Williams, Winn, and
  Zisserman]{everingham2010Pascal}
Mark Everingham, Luc Van~Gool, Christopher~KI Williams, John Winn, and Andrew
  Zisserman.
\newblock The pascal visual object classes (voc) challenge.
\newblock \emph{IJCV}, 88\penalty0 (2):\penalty0 303--338, 2010.

\bibitem[He and Zhang(2019)]{he2019multi}
Zhenwei He and Lei Zhang.
\newblock Multi-adversarial faster-rcnn for unrestricted object detection.
\newblock In \emph{ICCV}, 2019.

\bibitem[He and Zhang(2020)]{he2020domain}
Zhenwei He and Lei Zhang.
\newblock Domain adaptive object detection via asymmetric tri-way faster-rcnn.
\newblock In \emph{ECCV}, 2020.

\bibitem[Hsu et~al.(2020)Hsu, Tsai, Lin, and Yang]{hsu2020every}
Cheng-Chun Hsu, Yi-Hsuan Tsai, Yen-Yu Lin, and Ming-Hsuan Yang.
\newblock Every pixel matters: Center-aware feature alignment for domain
  adaptive object detector.
\newblock In \emph{ECCV}, 2020.

\bibitem[Inoue et~al.(2018)Inoue, Furuta, Yamasaki, and Aizawa]{inoue2018cross}
Naoto Inoue, Ryosuke Furuta, Toshihiko Yamasaki, and Kiyoharu Aizawa.
\newblock Cross-domain weakly-supervised object detection through progressive
  domain adaptation.
\newblock In \emph{CVPR}, 2018.

\bibitem[Jiang et~al.(2021)Jiang, Chen, Wang, and Long]{jiang2021decoupled}
Junguang Jiang, Baixu Chen, Jianmin Wang, and Mingsheng Long.
\newblock Decoupled adaptation for cross-domain object detection.
\newblock \emph{arXiv preprint arXiv:2110.02578}, 2021.

\bibitem[Johnson-Roberson et~al.(2016)Johnson-Roberson, Barto, Mehta, Sridhar,
  Rosaen, and Vasudevan]{johnson2016driving}
Matthew Johnson-Roberson, Charles Barto, Rounak Mehta, Sharath~Nittur Sridhar,
  Karl Rosaen, and Ram Vasudevan.
\newblock Driving in the matrix: Can virtual worlds replace human-generated
  annotations for real world tasks?
\newblock \emph{arXiv preprint arXiv:1610.01983}, 2016.

\bibitem[Kalluri et~al.(2022)Kalluri, Sharma, and Chandraker]{memsac}
Tarun Kalluri, Astuti Sharma, and Manmohan Chandraker.
\newblock {MemSAC:Memory Augmented Sample Consistency for Large Scale Domain
  Adaptation}.
\newblock In \emph{ECCV}, 2022.

\bibitem[Kim et~al.(2019)Kim, Jeong, Kim, Choi, and Kim]{kim2019diversify}
Taekyung Kim, Minki Jeong, Seunghyeon Kim, Seokeon Choi, and Changick Kim.
\newblock Diversify and match: A domain adaptive representation learning
  paradigm for object detection.
\newblock In \emph{CVPR}, 2019.

\bibitem[Li et~al.(2020)Li, Du, Zhang, Wen, Luo, Wu, and Zhu]{li2020spatial}
Congcong Li, Dawei Du, Libo Zhang, Longyin Wen, Tiejian Luo, Yanjun Wu, and
  Pengfei Zhu.
\newblock Spatial attention pyramid network for unsupervised domain adaptation.
\newblock In \emph{ECCV}, 2020.

\bibitem[Li et~al.(2022)Li, Dai, Ma, Liu, Chen, Wu, He, Kitani, and
  Vajda]{li2022cross}
Yu-Jhe Li, Xiaoliang Dai, Chih-Yao Ma, Yen-Cheng Liu, Kan Chen, Bichen Wu,
  Zijian He, Kris Kitani, and Peter Vajda.
\newblock Cross-domain adaptive teacher for object detection.
\newblock In \emph{CVPR}, 2022.

\bibitem[Lin et~al.(2014)Lin, Maire, Belongie, Hays, Perona, Ramanan,
  Doll{\'a}r, and Zitnick]{lin2014microsoft}
Tsung-Yi Lin, Michael Maire, Serge Belongie, James Hays, Pietro Perona, Deva
  Ramanan, Piotr Doll{\'a}r, and C~Lawrence Zitnick.
\newblock Microsoft coco: Common objects in context.
\newblock In \emph{ECCV}, 2014.

\bibitem[Na et~al.(2017)Na, Lee, Kim, and Kim]{na2017read}
Seil Na, Sangho Lee, Jisung Kim, and Gunhee Kim.
\newblock A read-write memory network for movie story understanding.
\newblock In \emph{ICCV}, 2017.

\bibitem[Oh et~al.(2019)Oh, Lee, Xu, and Kim]{oh2019video}
Seoung~Wug Oh, Joon-Young Lee, Ning Xu, and Seon~Joo Kim.
\newblock Video object segmentation using space-time memory networks.
\newblock In \emph{ICCV}, 2019.

\bibitem[Rezaeianaran et~al.(2021)Rezaeianaran, Shetty, Aljundi, Reino, Zhang,
  and Schiele]{rezaeianaran2021seeking}
Farzaneh Rezaeianaran, Rakshith Shetty, Rahaf Aljundi, Daniel~Olmeda Reino,
  Shanshan Zhang, and Bernt Schiele.
\newblock Seeking similarities over differences: Similarity-based domain
  alignment for adaptive object detection.
\newblock In \emph{ICCV}, 2021.

\bibitem[Rodriguez and Mikolajczyk(2019)]{rodriguez2019domain}
Adrian~Lopez Rodriguez and Krystian Mikolajczyk.
\newblock Domain adaptation for object detection via style consistency.
\newblock \emph{arXiv preprint arXiv:1911.10033}, 2019.

\bibitem[Saito et~al.(2019)Saito, Ushiku, Harada, and Saenko]{saito2019strong}
Kuniaki Saito, Yoshitaka Ushiku, Tatsuya Harada, and Kate Saenko.
\newblock Strong-weak distribution alignment for adaptive object detection.
\newblock In \emph{CVPR}, 2019.

\bibitem[Sakaridis et~al.(2018)Sakaridis, Dai, and
  Van~Gool]{sakaridis2018semantic}
Christos Sakaridis, Dengxin Dai, and Luc Van~Gool.
\newblock Semantic foggy scene understanding with synthetic data.
\newblock \emph{IJCV}, 126\penalty0 (9):\penalty0 973--992, 2018.

\bibitem[Shen et~al.(2019)Shen, Maheshwari, Yao, and Savvides]{shen2019scl}
Zhiqiang Shen, Harsh Maheshwari, Weichen Yao, and Marios Savvides.
\newblock Scl: Towards accurate domain adaptive object detection via gradient
  detach based stacked complementary losses.
\newblock \emph{arXiv}, 2019.

\bibitem[Sinha et~al.(2020)Sinha, Ohashi, and Nakamura]{sinha2020class}
Saptarshi Sinha, Hiroki Ohashi, and Katsuyuki Nakamura.
\newblock Class-wise difficulty-balanced loss for solving class-imbalance.
\newblock In \emph{ACCV}, 2020.

\bibitem[Su et~al.(2020)Su, Wang, Zeng, Tang, Chen, Qiu, and
  Wang]{su2020adapting}
Peng Su, Kun Wang, Xingyu Zeng, Shixiang Tang, Dapeng Chen, Di~Qiu, and
  Xiaogang Wang.
\newblock Adapting object detectors with conditional domain normalization.
\newblock In \emph{ECCV}, 2020.

\bibitem[Sukhbaatar et~al.(2015)Sukhbaatar, Weston, Fergus,
  et~al.]{sukhbaatar2015end}
Sainbayar Sukhbaatar, Jason Weston, Rob Fergus, et~al.
\newblock End-to-end memory networks.
\newblock \emph{NeurIPS}, 28, 2015.

\bibitem[Tian et~al.(2021)Tian, Zhang, Wang, Xiang, and Pan]{tian2021knowledge}
Kun Tian, Chenghao Zhang, Ying Wang, Shiming Xiang, and Chunhong Pan.
\newblock Knowledge mining and transferring for domain adaptive object
  detection.
\newblock In \emph{ICCV}, 2021.

\bibitem[Vs et~al.(2021)Vs, Gupta, Oza, Sindagi, and Patel]{vs2021mega}
Vibashan Vs, Vikram Gupta, Poojan Oza, Vishwanath~A Sindagi, and Vishal~M
  Patel.
\newblock Mega-cda: Memory guided attention for category-aware unsupervised
  domain adaptive object detection.
\newblock In \emph{CVPR}, 2021.

\bibitem[Weston et~al.(2014)Weston, Chopra, and Bordes]{weston2014memory}
Jason Weston, Sumit Chopra, and Antoine Bordes.
\newblock Memory networks.
\newblock \emph{arXiv preprint arXiv:1410.3916}, 2014.

\bibitem[Xie et~al.(2019)Xie, Yu, Wang, Wang, and Zhang]{xie2019multi}
Rongchang Xie, Fei Yu, Jiachao Wang, Yizhou Wang, and Li~Zhang.
\newblock Multi-level domain adaptive learning for cross-domain detection.
\newblock In \emph{Proceedings of the IEEE/CVF International Conference on
  Computer Vision Workshops}, 2019.

\bibitem[Xu et~al.(2020{\natexlab{a}})Xu, Zhao, Jin, and Wei]{xu2020exploring}
Chang-Dong Xu, Xing-Ran Zhao, Xin Jin, and Xiu-Shen Wei.
\newblock Exploring categorical regularization for domain adaptive object
  detection.
\newblock In \emph{CVPR}, 2020{\natexlab{a}}.

\bibitem[Xu et~al.(2020{\natexlab{b}})Xu, Wang, Ni, Tian, and
  Zhang]{xu2020cross}
Minghao Xu, Hang Wang, Bingbing Ni, Qi~Tian, and Wenjun Zhang.
\newblock Cross-domain detection via graph-induced prototype alignment.
\newblock In \emph{CVPR}, 2020{\natexlab{b}}.

\bibitem[Yang and Chan(2018)]{yang2018learning}
Tianyu Yang and Antoni~B Chan.
\newblock Learning dynamic memory networks for object tracking.
\newblock In \emph{ECCV}, 2018.

\bibitem[Zhang et~al.(2021)Zhang, Wang, and Mao]{zhang2021rpn}
Yixin Zhang, Zilei Wang, and Yushi Mao.
\newblock Rpn prototype alignment for domain adaptive object detector.
\newblock In \emph{CVPR}, 2021.

\bibitem[Zhao et~al.(2020)Zhao, Guo, Shen, and Ye]{zhao2020adaptive}
Zhen Zhao, Yuhong Guo, Haifeng Shen, and Jieping Ye.
\newblock Adaptive object detection with dual multi-label prediction.
\newblock In \emph{ECCV}, 2020.

\bibitem[Zheng et~al.(2020)Zheng, Huang, Liu, and Wang]{zheng2020cross}
Yangtao Zheng, Di~Huang, Songtao Liu, and Yunhong Wang.
\newblock Cross-domain object detection through coarse-to-fine feature
  adaptation.
\newblock In \emph{CVPR}, 2020.

\bibitem[Zhu et~al.(2019)Zhu, Pang, Yang, Shi, and Lin]{zhu2019adapting}
Xinge Zhu, Jiangmiao Pang, Ceyuan Yang, Jianping Shi, and Dahua Lin.
\newblock Adapting object detectors via selective cross-domain alignment.
\newblock In \emph{CVPR}, 2019.

\bibitem[Zhuang et~al.(2020)Zhuang, Han, Huang, and Scott]{zhuang2020ifan}
Chenfan Zhuang, Xintong Han, Weilin Huang, and Matthew Scott.
\newblock ifan: Image-instance full alignment networks for adaptive object
  detection.
\newblock In \emph{AAAI}, 2020.

\end{thebibliography}

\section{Supplementary Material}

\maketitle
\subsection{Additional Result: Pascal VOC$\xrightarrow{}$Clipart1k}\label{sec:voc_clip_result}
We conducted experiments on one more dissimilar domain using the Pascal VOC Dataset as the source domain and Clipart1k as the target domain. As shown in~\cref{tab:table1}, MILA achieves an impressive mAP of $49.9\%$, outperforming all its counterparts. In particular, exhibits significant improvements in average precision for various categories, including bicycle ($+4.8$), bird ($+1.2$), car ($+4.9$), table ($+2.5$), person ($+2.5$), and plant ($+5.3$). These enhancements exceed those of the second-best model by a large margin, highlighting MILA's superior adaptability to large domain gaps.


\begin{table*}[t]
 \footnotesize	
  \centering
  \addtolength{\tabcolsep}{-3.0pt}
  \scalebox{0.7}{
    \begin{tabular}{l|cccccccccccccccccccc|r}
      \toprule 
      Method & aero & bcyle & bird & boat & bottle & bus & car & cat & chair & cow & table & dog & hrs & m-bike & prsn & plnt & sheep & sofa & train & tv & mAP\\
      \midrule 
      Source & 23.0 & 39.6 & 20.1 & 23.6 & 25.7 & 42.6 & 25.2 & 0.9 & 41.2 & 25.6 & 23.7 & 11.2 & 28.2 & 49.5 & 45.2 & 46.9 & 9.1 & 22.3 & 38.9 & 31.5 & 28.8 \\
      \midrule 
      SCL~\cite{shen2019scl} & \textbf{44.7} & 50.0 & 33.6 & 27.4 & 42.2 & 55.6 & 38.3 & \textbf{19.2} & 37.9 & \textbf{69.0} & 30.1 & 26.3 & 34.4 & 67.3 & 61.0 & 47.9 & 21.4 & 26.3 & 50.1 & 47.3 & 41.5 \\
      SWDA~\cite{saito2019strong} & 26.2 & 48.5 & 32.6 & 33.7 & 38.5 & 54.3 & 37.1 & 18.6 & 34.8 & 58.3 & 17.0 & 12.5 & 33.8 & 65.5 & 61.6 & 52.0 & 9.3 & 24.9 & 54.1 & 49.1 & 38.1 \\
      DM~\cite{kim2019diversify} & 25.8 & 63.2 & 24.5 & \textbf{42.4} & 47.9 & 43.1 & 37.5 & 9.1&  47.0 & 46.7 & 26.8 & 24.9 & 48.1 & 78.7 & 63.0 & 45.0 & 21.3 & 36.1 & \textbf{52.3} & \textbf{53.4} & 41.8\\
      CRDA~\cite{xu2020exploring} & 28.7 & 55.3 & 31.8 &  26.0 & 40.1 & \textbf{63.6} & 36.6 &  9.4 & 38.7 & 49.3 & 17.6 & 14.1 & 33.3 & 74.3 & 61.3 & 46.3 & 22.3 & 24.3 & 49.1 & 44.3 & 38.3 \\
      HTCN~\cite{chen2020harmonizing} & 33.6 & 58.9 & 34.0 & 23.4 & 45.6 & 57.0 & 39.8 & 12.0 & 39.7 & 51.3 & 21.1 & 20.1 & 39.1 & 72.8 & 63.0 & 43.1 & 19.3 & 30.1 & 50.2 & 51.8 & 40.3\\
      UMT~\cite{deng2021unbiased} & 39.6 & 59.1 &  32.4 & 35.0 & 45.1 & 61.9 & 48.4 & 7.5 & 46.0 & 67.6 & 21.4 & \textbf{29.5} & 48.2 & 75.9 & 70.5 & 56.7 & 25.9 & 28.9 & 39.4 & 43.6 & 44.1 \\
      AT~\cite{li2022cross} & 31.1 & 75.2 & 32.9 & 35.5 & \textbf{60.5} & 44.8 & 56.6 & 3.3 & \textbf{60.0} & 56.4 & 41.5 & 14.7 & \textbf{61.6} & \textbf{85.7} & 76.5 & 57.2&  \textbf{31.6} & 35.9 & 47.7 & 46.2 & 47.7 \\
            
      \midrule 
      Ours & 28.3  & \textbf{80.0}  & \textbf{35.2} & 42.0 & 56.7 & 44.6 & \textbf{61.5} & 9.3 & 59.0 & 62.2 & \textbf{44.0} & 24.2 & 60.9 & 77.0 & \textbf{79.0} & \textbf{62.5} & 29.3 & \textbf{45.1} & 49.8 & 47.8 & \textbf{49.9}  \\    
      \midrule 
     Oracle & 33.3 & 47.6 & 43.1 & 38.0 & 24.5 & 82.0 & 57.4 & 22.9 &  48.4 & 49.2 & 37.9 & 46.4 & 41.1 & 54.0 & 73.7 & 39.5 & 36.7 & 19.1 &  53.2 & 52.9 & 45.0\\
      \bottomrule 
    \end{tabular}
    }
  \caption{Results on the Clipart1k test set for \textbf{Pascal VOC$\xrightarrow{}$Clipart1k} adaptation (ResNet-101). 
  }
 \label{tab:table1}
\end{table*}

\begin{table}[htp]
 \footnotesize	
  \centering
  \addtolength{\tabcolsep}{-3.0pt}
    \begin{tabular}{l|cccccc|r}
      \toprule 
      Method & bicycle & bird & car & cat & dog & person & mAP\\
      \midrule 
      Source & 84.2 & 44.5 & 53.0 & 24.9 & 18.8 & 56.3 & 46.9 \\
       \midrule 
      AT~\cite{li2022cross} & 94.3 & 57.2 & 57.2 & 34.2 & 36.9 & 78.5 & 59.7\\
      AT$^\dagger$~\cite{li2022cross} & 93.6 & 56.1 & \textbf{58.9} & 37.3 & 39.6 & 73.8 & 59.9\\
      \midrule 
      Ours & \textbf{97.4} & \textbf{59.0} & 58.3 & \textbf{40.6} & \textbf{47.8} & \textbf{79.3} & \textbf{63.7}\\
      \midrule 
      Oracle & 51.8 & 49.7 & 42.5 & 38.7 & 52.1 & 68.6 & 50.6\\
      \bottomrule 
    \end{tabular}
  \caption{Results on the Watercolor2k test set for \textbf{PASCAL VOC $\xrightarrow{}$ Watercolor2k} adaptation (ResNet-101). $\dagger$ represents the results copied from the origin paper \cite{li2022cross}.}
   \label{tab:table11_}
\end{table}

\begin{table*}[htp]
 \footnotesize	
  \centering
  \addtolength{\tabcolsep}{-4.0pt}
  \scalebox{0.75}{
    \begin{tabular}{l|cccccccccccccccccccc|r}
      \toprule 
      Method & aero & bcyle & bird & boat & bottle & bus & car & cat & chair & cow & table & dog & hrs & m-bike & prsn & plnt & sheep & sofa & train & tv & mAP\\
      \midrule 
      Source & 23.0 & 39.6 & 20.1 & 23.6 & 25.7 & 42.6 & 25.2 & 0.9 & 41.2 & 25.6 & 23.7 & 11.2 & 28.2 & 49.5 & 45.2 & 46.9 & 9.1 & 22.3 & 38.9 & 31.5 & 28.8 \\
       \midrule 
      AT~\cite{li2022cross} & 31.1 & 75.2 & 32.9 & 35.5 & \textbf{60.5} & 44.8 & 56.6 & 3.3 & \textbf{60.0} & 56.4 & 41.5 & 14.7 & \textbf{61.6} & \textbf{85.7} & 76.5 & 57.2&  \textbf{31.6} & 35.9 & 47.7 & 46.2 & 47.7 \\
      AT$^\dagger$~\cite{li2022cross} & 33.8 & 60.9 & \textbf{38.6} & \textbf{49.4} & 52.4 & 53.9 & 56.7 & 7.5 & 52.8 & 63.5 & 34.0 & 25.0 & \textbf{62.2} & 72.1 & 77.2 & 57.7 & 27.2 & \textbf{52.0} & \textbf{55.7} & \textbf{54.1} & 49.3 \\
            
      \midrule 
      Ours & 28.3  & \textbf{80.0}  & \textbf{35.2} & 42.0 & 56.7 & 44.6 & \textbf{61.5} & 9.3 & 59.0 & 62.2 & \textbf{44.0} & 24.2 & 60.9 & 77.0 & \textbf{79.0} & \textbf{62.5} & 29.3 & \textbf{45.1} & 49.8 & 47.8 & \textbf{49.9}  \\    
      \midrule 
     Oracle & 33.3 & 47.6 & 43.1 & 38.0 & 24.5 & 82.0 & 57.4 & 22.9 &  48.4 & 49.2 & 37.9 & 46.4 & 41.1 & 54.0 & 73.7 & 39.5 & 36.7 & 19.1 &  53.2 & 52.9 & 45.0\\
      \bottomrule 
    \end{tabular}
  }
  \caption{Results on the Clipart1k test set for \textbf{PASCAL VOC $\xrightarrow{}$ Clipart1k} adaptation (ResNet-101). $\dagger$ represents the results copied from the origin paper \cite{li2022cross}.
  }
 \label{tab:table22_}
\end{table*}

\begin{table*}
 \footnotesize	
  \centering
   \addtolength{\tabcolsep}{-0.5pt}
  \scalebox{0.75}{
    \begin{tabular*}{\textwidth}{@{\extracolsep{\fill}\quad} l|cccccccc|r}
      \toprule 
      Method & bus & bicycle & car & mcycle & person & rider & train & truck & mAP\\
      \midrule 
      Source (F-RCNN) & 20.1 & 31.9 & 39.6 & 16.9 & 29.0 & 37.2 & 5.2 & 8.1 & 23.5\\
       \midrule 
      AT~\cite{li2022cross} & 60.0 & 49.0 & 63.6 & 38.8 & 45.0 & 53.9 & 45.1 & 33.9 & 49.0\\
      AT$^\dagger$~\cite{li2022cross} & 56.3 & \textbf{51.9} & 64.2 & 38.5 & 45.5 & \textbf{55.1} & \textbf{54.3} & \textbf{35.0} & \textbf{50.9}\\
      \midrule 
      Ours & \textbf{61.4}  & \textbf{51.5} & \textbf{64.8} & \textbf{39.7} & \textbf{45.6} & 52.8 & 54.1 & 34.7 & \textbf{50.6}\\    
      \midrule 
       Oracle & 50.3 & 40.7 & 61.3 & 32.5 & 43.1 & 49.8 & 35.1 & 28.6 & 42.7\\
      \bottomrule 
    \end{tabular*}
    }
   \caption{Results on the Foggy Cityscapes test set for \textbf{Cityscapes $\xrightarrow{}$ Foggy Cityscapes} adaptation (VGG-16). $\dagger$ represents the results copied from the origin paper \cite{li2022cross}}.
    \label{tab:table33_}
\end{table*}

\begin{table}[t!]
\footnotesize
\centering
\begin{tabular*}{0.45 \textwidth}{@{\extracolsep{\fill}\quad}l|ccc|c}
\toprule
\# & $\mathcal{L}_{Unsup}$ & $\mathcal{L}_{Dis}$ & $\mathcal{L}_{Ins}$ & mAP \\
\midrule
1 & \checkmark & \checkmark & \checkmark & $\mathbf{63.7}\hspace{1mm} (-0.0)$ \\
2 &\checkmark & \checkmark & \xmark & $59.7\hspace{1mm} (-4.0)$ \\
3 & \checkmark & \xmark & \checkmark & $60.4\hspace{1mm} (-3.3)$ \\
4 &\xmark & \checkmark & \checkmark & $55.5 \hspace{1mm}(-8.2)$ \\
\bottomrule
\end{tabular*}
 \caption{Effect of different loss components ($\mathcal{L}_{Unsup}$, $\mathcal{L}_{Dis}$ and $\mathcal{L}_{Ins}$) on domain adaptation using MILA. While \xmark  denotes that the loss component is switched off, \checkmark denotes that the component is on and used. $\mathcal{L}_{Unsup}$, $\mathcal{L}_{Dis}$ and $\mathcal{L}_{Ins}$ can be switched off simply by setting $\lambda_1, \lambda_2$ and $\lambda_3$ to 0 respectively.}
    \label{tab:table44_}
\end{table}

\subsection{Comparisons with Reported Results in~\cite{li2022cross}}\label{sec:orig_results}
As mentioned in the main paper, we implemented our method on top of the official code of~\cite{li2022cross}. However, using the official code, we could not reproduce the results as reported in \cite{li2022cross}. Note that this issue has been previously reported by others in the official repository.
\cref{tab:table11_,tab:table22_,tab:table33_} show that though the reported results in \cite{li2022cross} are slightly higher than the reproduced ones, our model still outperforms the reported results on Watercolor2k ($+3.8$) and Clipart1k ($+0.6$) adaptations.

\subsection{Comparison with MeGA-CDA~\cite{vs2021mega}}\label{sec:ablation_}


In the main paper, we briefly mentioned the main differences between MILA and MeGA-CDA in~\cref{sec:related}. However, in this section, we aim to provide further experiments and evidence to thoroughly discuss and support these differences.

\paragraph{Missing Pairs for Alignment in MeGA-CDA.} Both MILA and MeGA-CDA utilize memory to store source proposals, but they differ in how these stored proposals are employed for domain alignment. MeGA-CDA focuses on achieving category-to-category (C2C) alignment by highlighting category-specific regions within the backbone features of source-target mini-batch pairs.
In contrast, MILA aims not only to align categories but also to identify the most visually similar pair of the same category for alignment purposes for each object proposal.
When it comes to MeGA-CDA, if the matching category is absent in the source-target pair, object regions remain unaligned or misaligned, with an erroneously highlighted region in the image. As depicted in Figure 1(b), 
the gray (MeGA-CDA) dots are sparser compared to the orange (MILA) dots. This indicates that many target instances lack corresponding source instances within the mini-batch, which existing methods overlook in the alignment process. \textbf{Specifically, MeGA-CDA fails to locate matching source instances of the same category for a significant proportion of target instances (39\%).} In contrast, MILA guarantees the retrieval of a visually similar source instance of the same category for every target instance, thanks to its memory storage and retrieval strategy.

\paragraph{Lower Intra-class Variance in Memory of MeGA-CDA.} MeGA-CDA regularizes memory to make it more compact, which reduces intra-class variance, whereas MILA preserves intra-class variance and makes use of it to retrieve the most similar pair for alignment. This enables the model to focus on its primary task of domain alignment when the candidates for alignment are of the same category and visually very similar. We experimentally evaluated the intra-class variance of storage in MILA and MeGA-CDA using Frechet Inception Distance (FID), a popular measure of diversity in GAN literature, on the Pascal VOC dataset. The average FID score of our storage in MILA is 143.1, whereas MeGA-CDA scores 73.2, suggesting that the intra-class variance in MILA is much higher than in MeGA-CDA.

\begin{table}[t]
 \footnotesize	
  \centering
    \begin{tabular*}{0.45\textwidth}{@{\extracolsep{\fill}\quad} l|ccccc}
      \toprule 
      $\lambda_3$ & 0.0 & 0.05 & 0.10 & 0.15  & 0.20  \\
      \midrule 
      mAP & 59.7 & 62.0 & \textbf{63.7} & 60.0 & 57.3  \\
      \bottomrule 
    \end{tabular*}
  \caption{
  Effect of $\mathcal{L}_{Ins}$ on performance of MILA. We vary $\lambda_3$ in [0,0.2] to control the impact of $\mathcal{L}_{Ins}$. 
  }
    \label{tab:table22}
 
\end{table}

\begin{table}[t!]
 \footnotesize	
  \centering
    \begin{tabular*}{0.45 \textwidth}{@{\extracolsep{\fill}\quad} l|cccc}
      \toprule 
      $K$ & 1 & 10 & 30 & 100 \\
      \midrule 
      mAP & \textbf{63.7} & 60.8 & 59.0  & 56.8\\
      \bottomrule 
    \end{tabular*}
  \caption{Effect of varying $K$. Note that we retrieve top-$K$ similar source instance features from memory for a target instance.
  }
    \label{tab:table33}
\end{table}

\begin{table}[t]
 \footnotesize	
  \centering
  \addtolength{\tabcolsep}{-3.0pt}
   \begin{tabular*}{0.6 \textwidth}{@{\extracolsep{\fill}\quad} l|cccc}
      \toprule 
      storage method & $\gamma = 0.1$ & $\gamma = 0.2$ & $\gamma = 0.3$ & $\gamma=1$\\
      \midrule 
      category-imbalanced & 56.7 & 59.2 & 60.2 & 63.7\\
      category-balanced & $\mathbf{58.6}$ & 
      $\mathbf{60.7}$ & 
      $\mathbf{61.2}$ &  --\\
      \bottomrule 
    \end{tabular*}
 \caption{Comparison of mAP obtained by category-balanced and category-imbalanced memory storage methods. For $\gamma=1$, it is not possible to obtain enough source features for the minority categories to maintain a category-balanced memory.}
    \label{tab:table55_}
\end{table}

\subsection{Hyperparameter Analysis}\label{sec:hyperparameter_analysis}
In this section, we delve deeper into our approach by examining the impact of various hyperparameters. We thoroughly analyze the effects using the Pascal VOC$\xrightarrow{}$Watercolor2k dataset.

\noindent\textbf{Significance of different loss components.} 
%
As shown in~\cref{eq:overall_loss}, in addition to source-supervised loss $\mathcal{L}_{Sup}$, the overall training loss $\mathcal{L}$ for MILA is comprised of unsupervised loss $\mathcal{L}_{Unsup}$ from target pseudo-labels, discriminator loss $\mathcal{L}_{Dis}$ for image level alignment and our proposed memory-based instance alignment loss $\mathcal{L}_{Ins}$. 
Note that $\mathcal{L}_{Dis}$ and $\mathcal{L}_{Ins}$ specifically aim to adapt the model from source to target domain whereas $\mathcal{L}_{Unsup}$ aims to optimize the model with target pseudo-labels generated as in~\cite{li2022cross}.
In \cref{tab:table44_}, we analyse the effect of each of these loss components on MILA. As can be seen, all the three components are crucial for the effectiveness for MILA. 
Especially for domain alignment losses ($\mathcal{L}_{Dis}$ and $\mathcal{L}_{Ins}$), we observe that the performance drops more when $\mathcal{L}_{Ins}$ is not used (\#1 vs. \#2) than compared to when $\mathcal{L}_{Dis}$ is not used (\#1 vs. \#3), which suggests that instance level adaptation is crucial for domain alignment.  

%
%
\paragraph{Effect of instance-level alignment weight ( $\lambda_3$).}
To further analyze the importance of MILA's instance-level alignment module, we varied $\lambda_3$, which determines the weight for instance-level alignment in the training process and investigated its influence on the model’s performance.
A large value means we are giving more importance to instance-level adaptation.
If $\lambda_3$ is 0, the model performs image-level adaptation only. As shown in~\cref{tab:table22}, a high or low value of $\lambda_3$ can degrade the performance, and the peak accuracy is achieved when $\lambda_3$ is 0.1. 
Recalling that the default value of $\lambda_2$ in our experiment is 0.1, this result means that it is good to penalize the misalignment in image level and instance level to the same extent.
%
%

\paragraph{Effect of the number of retrieved memory pairs ($K$).}
Parameter $K$ determines how many matching instances to retrieve from memory based on the cosine similarity score. 
As shown in~\cref{tab:table33}, the peak accuracy is achieved when $K$ is 1, i.e., only the most similar source instance is retrieved and aligned with the target instance. This validates our claim that, for most similar instances of two domains, our model can focus on the primary task of adapting the two domains without being disturbed by their intra-category variations.
%

\noindent\textbf{Comparison of different memory storage techniques.}
We aim to understand the impact of memory size on MILA's performance by selectively storing only a fraction of the total instances for each category.
To control the number of instances stored, we utilize a hyperparameter denoted as $\gamma$. Specifically, the number of stored instance features, denoted as $M(c)$, for category $c$ in memory is determined as:
\begin{equation}
    |M(c)| = \ceil[\big]{\gamma \sum_{i=1}^{N^S} \sum_{j=1}^{K^S_i} \mathbf{1}(c^S_{i,j} = c)},
\label{eqn:storage}
\end{equation}
Here, $\gamma$ ranges from 0 to 1. In~\cref{eqn:storage}, the underlying concept is that we add instance features to memory for class $c$ as long as the number of stored instance features $|M(c)|$ is less than $\gamma$ times the total number of instances for class $c$ in the source dataset $D_S$. This results in a \textbf{category-imbalanced} memory, where the number of stored instances varies across categories.

Alternatively, we can employ another approach where an equal number of source instance features per category are stored in memory. In this case, the number of stored instance features $|M(c)|$ for category $c$ is given by:
\begin{equation}
    |M(c)| = \frac{\gamma \sum_{i=1}^{N^S} K^S_i}{C},
\end{equation}
This storage technique creates a \textbf{category-balanced} memory, which helps alleviate the impact of class imbalance in the domain alignment.
In \cref{tab:table55_}, we compare these two storage techniques. In both methods, the mean average precision (mAP) increases as $\gamma$ increases. Interestingly, for a given $\gamma$, the category-balanced storage method outperforms the category-imbalanced one. We hypothesize that this is due to the imbalance-mitigating effect of the former approach.



\subsection{Limitations}
As shown in
and \cref{tab:table55_}, 
the performance of MILA increases monotonically with the memory storage size. However, creating the memory with high storage size and dynamically updating it increases the training time and memory consumption. This is a trade-off between memory usage and performance, which is left to be solved in future work.




\end{document}